\documentclass[sigconf,balance=false,tikz, anonymous=false,review=false]{acmart}

\usepackage{smartdiagram}
\usepackage{graphicx}
\usepackage{booktabs}
\usepackage{tikz}
\usetikzlibrary{shapes,arrows}
\usetikzlibrary{positioning}
\usepackage{caption}  
\usepackage[position=top]{subfig}
\usepackage{scalefnt}
\usepackage{amsmath}
\usepackage{amsthm}
\usepackage{varwidth}
\usepackage{scalerel}
\usepackage{algorithm}
\usepackage{algpseudocode}
\usepackage{multicol}
\usepackage{multirow}
\usepackage{float}
\usepackage{adjustbox}
\usepackage{array}
\usepackage{bbding}
\usepackage{pifont}
\usepackage{hyperref}

\newcolumntype{R}[2]{%
    >{\adjustbox{angle=#1,lap=\width-(#2)}\bgroup}%
    l%
    <{\egroup}%
}
\newcommand*\rot{\multicolumn{1}{R{45}{1em}}}

\algrenewcommand\textproc{}

\makeatletter
\newcommand*{\algrule}[1][\algorithmicindent]{\makebox[#1][l]{\hspace*{.5em}\vrule height .75\baselineskip depth .25\baselineskip}}%

\newcount\ALG@printindent@tempcnta
\def\ALG@printindent{%
    \ifnum \theALG@nested>0
        \ifx\ALG@text\ALG@x@notext
            \addvspace{-3pt}
        \else
            \unskip
            \ALG@printindent@tempcnta=1
            \loop
                \algrule[\csname ALG@ind@\the\ALG@printindent@tempcnta\endcsname]%
                \advance \ALG@printindent@tempcnta 1
            \ifnum \ALG@printindent@tempcnta<\numexpr\theALG@nested+1\relax
            \repeat
        \fi
    \fi
    }%
    
\algnewcommand\algorithmicforeach{\textbf{for each}}
\algdef{S}[FOR]{ForEach}[1]{\algorithmicforeach\ #1\ \algorithmicdo}

\algblockdefx[Foreach]{Foreach}{EndForeach}[1]{\textbf{foreach} #1 \textbf{do}}{\textbf{end foreach}}

\algnewcommand{\algorithmicand}{\textbf{ and }}
\algnewcommand{\algorithmicor}{\textbf{ or }}
\algnewcommand{\OR}{\algorithmicor}
\algnewcommand{\AND}{\algorithmicand}
\algnewcommand{\var}{\texttt}
\let\oldReturn\Return
\renewcommand{\Return}{\State\oldReturn}

\algnewcommand{\LineComment}[1]{\State \(\triangleright\) \textit{\color{blue} \scriptsize #1}}

\tikzset{
  treenode/.style = {align=center, inner sep=0pt, text centered,
    font=\sffamily},
  arn_n/.style = {treenode, circle, black, font=\sffamily, draw=black,
    fill=white, text width=1.5em},
  arn_r/.style = {treenode, circle, white, draw=red, fill=red,
    text width=1.5em},
  arn_b/.style = {treenode, circle, white, draw=blue, fill=blue,
    text width=1.5em},
  arn_br/.style = {treenode, circle, black, draw=brown, fill=brown!40,
    text width=1.5em},
  arn_v/.style = {treenode, circle, black, draw=blue, fill=violet!30,
    text width=1.5em},
}

\tikzstyle{decision} = [diamond, draw, fill=orange!20, 
    text width=6em, text badly centered,node distance=2cm, inner sep=0pt]
\tikzstyle{block} = [rectangle, draw, fill=blue!20, 
    text width=6em, text centered, rounded corners, node distance=1.5 cm, minimum height=2em]
\tikzstyle{block2} = [rectangle, draw, fill=red!20, 
    text width=6em, text centered, rounded corners, node distance=1.5 cm, minimum height=2em]
\tikzstyle{line} = [draw, -latex']
\tikzstyle{cloud} = [draw, ellipse,fill=red!20, node distance=3cm,
    minimum height=2em]

\AtBeginDocument{%
  \providecommand\BibTeX{{%
    \normalfont B\kern-0.5em{\scshape i\kern-0.25em b}\kern-0.8em\TeX}}}

\setcopyright{acmcopyright}
\copyrightyear{2024}
\acmYear{2024}
\setcopyright{acmlicensed}\acmConference[GECCO '24]{Genetic and
Evolutionary Computation Conference}{July 14--18, 2024}{Melbourne, VIC,
Australia}
\acmBooktitle{Genetic and Evolutionary Computation Conference (GECCO '24),
July 14--18, 2024, Melbourne, VIC, Australia}
\acmDOI{10.1145/3638529.3654092}
\acmISBN{979-8-4007-0494-9/24/07}
\acmConference[GECCO '24]{Genetic and Evolutionary Computation Conference}{July 14 - 18, 2024}{Melbourne, Australia}
\acmPrice{}

\author{Roman Kalkreuth}
\affiliation{%
    \institution{CNRS, LIP6, Sorbonne Universit\'e}
    \city{Paris}
    \country{France}}
\email{roman.kalkreuth@lip6.fr}
\orcid{0000-0003-1449-5131}

\author{Thomas B{\"a}ck}
\affiliation{%
      \institution{LIACS, Leiden University}
      \streetaddress{Niels Bohrweg 1}
      \city{Leiden}
      \postcode{2333}
      \country{The Netherlands}}
\email{t.h.w.baeck@liacs.leidenuniv.nl}
\orcid{0000-0001-6768-1478}

\theoremstyle{definition}
\newtheorem{definition}{Definition}[section]

\begin{document}

\title{CGP++ : A Modern C++ Implementation of Cartesian Genetic Programming}



\begin{abstract}
The reference implementation of Cartesian Genetic Programming (CGP) was written in the C programming language. C inherently follows a procedural programming paradigm, which entails challenges in providing a reusable and scalable implementation model for complex structures and methods. Moreover, due to the limiting factors of C, the reference implementation of CGP does not provide a generic framework and is therefore restricted to a set of predefined evaluation types. Besides the reference implementation, we also observe that other existing implementations are limited with respect to the features provided. In this work, we therefore propose the first version of a modern C++ implementation of CGP that pursues object-oriented design and generic programming paradigm to provide an efficient implementation model that can facilitate the discovery of new problem domains and the implementation of complex advanced methods that have been proposed for CGP over time. With the proposal of our new implementation, we aim to generally promote interpretability, accessibility and reproducibility in the field of CGP.
\end{abstract}



\maketitle

\section{Introduction}
\label{sec:intro}

Cartesian Genetic Programming (CGP) can be considered a well-established graph-based representation model of GP. The first pioneering work towards CGP was done by Miller, Thompson, Kalganova, and Fogarty~\cite{mi-th-fo-97a,ka-97a,miller:1999:ACGP} by the introduction of a graph encoding model based on a two-dimensional array of functional nodes. CGP can be considered an extension to the traditional tree-based representation model of GP since it enables many applications in problem domains that are well-suited for graph-based representations such as circuit design~\cite{DBLP:conf/ddecs/FiserSVS10, Sekanina2011}, neural architecture search~\cite{10.1162/evco_a_00253, 10.1162/evco_a_00253} and image processing~\cite{naturecommcgp}. Miller officially introduced CGP over 20 years ago~\cite{DBLP:conf/eurogp/MillerT00} and provided a reference implementation written in the C programming language. Since then, several implementations have been provided in other popular programming languages such as Java or Python, which follow modern programming paradigms. Besides implementations, several sophisticated methods for CGP have been proposed over time, and the significance of various developments has been recently surveyed and discussed in the context of the status and future of CGP~\cite{DBLP:journals/gpem/Miller20}. Miller's reference implementation is based on the procedural programming paradigm, which naturally entails challenges and limitations to provide a flexible, reusable and generic architecture that can facilitate the implementation of complex methods and their corresponding structures. Moreover, Julian F. Miller passed away in 2022~\cite{10.1162/artl_a_00371}\footnote{\url{http://www.evostar.org/2022/julian-francis-miller/}} and his website\footnote{\url{http://www.cartesiangp.co.uk/}} which served as a resource for his original implementation, disappeared shortly after his death for unknown reason to the authors of this paper. The above-described points and circumstances motivates our work, in which we present a modern implementation of CGP written in C++ called \texttt{CGP++}. Our implementation builds upon paradigms and methodologies commonly associated with the modern interpretation of the C++ programming language, such as generic programming. Since C++ has a reputation for providing excellent performance while representing a high-level object-oriented language that offers many features for generic programming, we feel that C++ is a suitable choice for a modern and contemporary implementation of CGP. This paper is structured as follows: In Section~\ref{sec:related_work} we describe GP and CGP and address major problem domains in these fields.   Section~\ref{sec:implementations} surveys existing implementations of CGP that have been proposed for various programming languages. In Section~\ref{sec:proposal} we introduce our new implementation by presenting key features and addressing relevant implementation details. Section~\ref{sec:overview} gives an overview of the architecture and workflow of \texttt{CGP++}. In Section~\ref{sec:comparison} we compare our implementation to the implementations that have been addressed in this paper. Section~\ref{sec:discussion} discusses the potential role of \texttt{CGP++} in the ecosystem of CGP implementations and addresses prospects as well as challenges of enhancements that will be considered by future work. Finally, Section~\ref{sec:conclusions} concludes our work.


\section{Related work} 
\label{sec:related_work}

\subsection{Genetic Programming} 

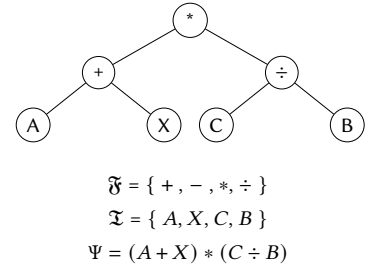
\begin{figure}
\scalebox{0.58}{
\scalefont{1.5}
\begin{tikzpicture}[level distance = 1.2cm, level/.style={sibling distance={6cm/max(2,#1)}}] 
\node [arn_n] {*}
    child{ node [arn_n, xshift=-6mm] {+} 
            child{ node [arn_n] {A}}
            child{ node [arn_n] {X}}                            
    }
      child{ node [arn_n, xshift=6mm] {$\div$} 
            child{ node [arn_n] {C}}
            child{ node [arn_n] {B}}                            
    }
;
\end{tikzpicture}
} \\ \vspace{3mm}
\small 
$\mathfrak{F}$ = $\lbrace$ $+$ , $-$ , $*$, $\div$ $\rbrace$ \\ \vspace{1mm}
$\mathfrak{T}$ = $\lbrace$ $A$, $X$, $C$, $B$ $\rbrace$ \\ \vspace{1mm}
$\Psi = (A + X) * (C \div  B) $
\caption{Example of a parse tree as used in conventional GP. A parse tree can be considered a composition of elements taken from the function set $\mathcal{F}$ and terminal set $\mathcal{T}$. The decoding of the parse tree in the example leads to the symbolic expression $\Psi$. }
\label{img:parse_tree}
\end{figure}

In the wider taxonomy of heuristics, Genetic Programming (GP) can be considered an evolutionary-inspired search heuristic that enables the synthesis of computer programs for problem-solving. The fundamental paradigm of GP aims at evolving a population of candidate \emph{computer programs} towards an algorithmic solution of a predefined problem. GP transforms candidate genetic programs (Definition~\ref{definition_gpr}), that are traditionally represented as parse-trees, iteratively from generation to generation into new populations of programs with (hopefully) better fitness. However, since GP is a stochastic optimization process, obtaining the optimal solution is consequently not guaranteed. A formal definition of GP is provided in Definition~\ref{definition_gp}. GP traditionally uses a parse-tree representation model that is inspired by LISP S-expressions. An example of a parse tree is illustrated in Figure~\ref{img:parse_tree}. In addition to the conventional (tree-based) GP, GP is also used with linear sequence representations~\cite{ieee94:perkis,Openshaw94buildingnew}, graph-based representation models~\cite{Poli1996,miller:1999:ACGP}, or grammar-based representations~\cite{ryan:1998:geepal}. 

\begin{definition}[Genetic Program]
A genetic program $\mathfrak{P}$  is an element of  $ \mathfrak{T} \times \mathfrak{F}  \times \mathfrak{E} $:
\begin{itemize}
\item $\mathfrak{F}$ is a finite non-empty set of functions \\ 
\item $\mathfrak{T}$ is a finite non-empty set of terminals \\ 
\item $\mathfrak{E}$ is a finite non-empty set of edges \\ 
\end{itemize}
Let $\phi$: $\mathcal{P} \mapsto \Psi$ be a decode function which maps $\mathcal{P}$ to a phenotype $\Psi$	  
\label{definition_gpr}
\end{definition}

\begin{definition}[Genetic Programming]
Genetic Programming is an evolutionary algorithm-based method for the automatic derivation of computer programs. 
Let $\mathfrak{B}_\mu^{(g)}$ be a population of $\mu$ individuals and let $\mathfrak{B}_\mu^{(g+1)}$ be the population of the following generation: 
\begin{itemize}
\item Each individual is represented with a genetic program and a fitness value.
\item Genetic Programming transforms $\mathfrak{B}_\mu^{(g)} \mapsto \mathfrak{B}_\mu^{(g+1)}$ by the adaptation of selection, recombination and mutation. 
\end{itemize}
\label{definition_gp}
\end{definition}

\subsection{Cartesian Genetic Programming} 
\label{sec:cgp}

CGP can be considered an extension of conventional tree-based GP since it represents a genetic program as an acyclic and directed graph, and trees as data structures naturally entail combinational limitations. A genetic program is encoded in the genotype of an individual and is decoded to its corresponding phenotype before evaluation. Originally, the programs were represented with a rectangular $n_\textnormal{r}$  $\times$ $n_\textnormal{c}$ grid of nodes. However, later work focused on a representation with merely one row. A formal definition of a cartesian genetic program (CP) is given in Definition~\ref{definition_cp}. In CGP, \textit{function nodes} are used to execute functions, defined in the function set, on the input values.
The decoding routine distinguishes between groups of genes, and each group represents a node of the graph, except the last one, which refers to the outputs. Two types of genes are used to encode a node: 1) the function gene, that indexes the function number in the function set and 2) the connection genes, which index the inputs of the node. The number of connection genes varies based on the on the predefined maximum arity $n_\textnormal{a}$ of the function set. The decoding of function nodes is embedded in a backward search that is performed for all output genes. 

\begin{figure}
\centering
\includegraphics[scale=0.85]{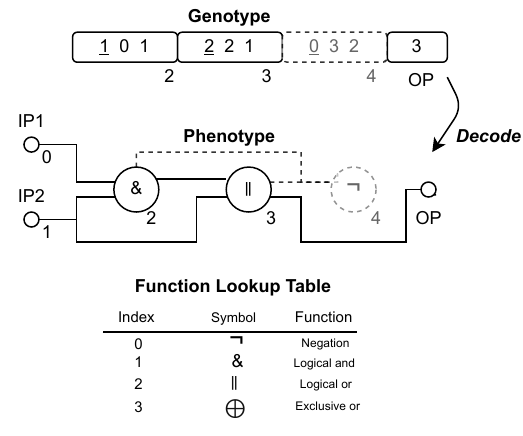}
\caption{Illustration of the decoding procedure of the CGP genotype to the corresponding phenotype. The identifiers IP1 and IP2 refer to the two input nodes with node index $0$ and $1$. The identifier OP stands for the output node of the graph.}
\label{cgp_decoding}
\end{figure}

The backward search is illustrated in Figure~\ref{cgp_decoding} for the commonly used single-row integer representation, which starts from the program output and processes all linked nodes in the genotype until the inputs are reached. Consequently, only active nodes are processed during evaluation. The genotype illustrated in Figure~\ref{cgp_decoding} is grouped whereby the first (underlined) gene of each group refers to the function number in the function set. In contrast, the non-underlined genes which refer to the respective input connections of the node. Function nodes, that are not linked in the genotype, remain inactive and are visualized in gray color as well as dashed lines. A parameter called levels-back $l$ is commonly used to control the connectivity of the graph by constraining the node index from which a function or output node can get its inputs. 
 

\begin{definition}[Cartesian Genetic Program]
A cartesian genetic program is an element of the Cartesian product $\mathfrak{N_\textnormal{i}}  \times \mathfrak{N_\textnormal{f}}  \times \mathfrak{N_\textnormal{o}}  \times \mathfrak{F}: $ 
\begin{itemize}\setlength\itemsep{-0.5em}
\setlength\itemsep{-0.7em}
\item $\mathfrak{N_\textnormal{i}}$ is a finite non-empty set of input nodes \\ 
\item $\mathfrak{N_\textnormal{f}}$ is a finite set of function nodes \\ 
\item $\mathfrak{N_\textnormal{o}}$ is a finite non-empty set of output nodes \\ 
\item $\mathfrak{F}$ is a finite non-empty set of functions \\  
\end{itemize}  
\label{definition_cp}
\end{definition}


\begin{algorithm}
\caption{Exemplification of the (1+$\lambda$)-EA variant with neutral genetic drift}
\begin{algorithmic}[1]
\small
\State initialize($\mathcal{P}$) \Comment{Initialize parent individual}
\Repeat \Comment{Until termination criteria not triggered}
\State $\mathcal{Q} \gets \mathnormal{breed(\mathcal{P})}$ \Comment{Breed $\lambda$ offspring by mutation}

\State Evaluate($\mathcal{Q}$) \Comment{Evaluate the fitness of the offspring}

\State $\mathcal{Q}^\mathnormal{+} \gets \mathnormal{best(\mathcal{Q},\mathcal{P})}$ \Comment{Get individuals which have better fitness as the parent}
\State $\mathcal{Q}^\mathnormal{=} \gets \mathnormal{equal(\mathcal{Q},\mathcal{P})}$ \Comment{Get individuals which have the same fitness as the parent}
\LineComment{If there exist individuals with better fitness}
\If {$|\mathcal{Q}^\mathnormal{+}| > 0$} 
\LineComment{Choose one individual from $\mathcal{Q}^\mathnormal{+}$ uniformly at random}
\State $\mathcal{P} \gets \mathcal{Q}^\mathnormal{+}$[r], $r \sim U$[0, $|\mathcal{Q}^\mathnormal{+}| - 1$] 
\LineComment{Otherwise, if there exist individuals with equal fitness }
\ElsIf {$|\mathcal{Q}^\mathnormal{=}| > 0$}
\LineComment{Choose one individual from $\mathcal{Q}^\mathnormal{=}$ uniformly at random}
\State $\mathcal{P} \gets \mathcal{Q}^\mathnormal{=}$[r], $r \sim U$[0, $|\mathcal{Q}^\mathnormal{=}| - 1$] 
\EndIf
\Until {$\mathcal{P}$ \textit{meets termination criterion}} 
\State \textbf{return} $\mathcal{P}$ \Comment{}
\end{algorithmic}
\label{cgp_algorithm}
\end{algorithm}

The number of inputs $n_\textnormal{i}$, outputs $n_\textnormal{o}$, and the length of the genotype remain static during a run of CGP. Therefore, each candidate program can be represented with $n_\textnormal{r} * n_\textnormal{c} * (n_\textnormal{a} +1) + n_\textnormal{o}$ integers. However, although the length of the genotype is static, the length of the corresponding phenotype can vary during a run, which enables a certain degree of flexibility of the CGP representation model. CGP is commonly used with a ($1+\lambda$) evolutionary algorithm (EA) and a selection strategy called \textit{neutrality}, which is based on the idea that the adaption of a neutral genetic drift mechanism (NGD) can contribute significantly to the escape from local optima. NGD is implemented by extending the classical selection mechanism in such a way that individuals which have equal fitness as the normally selected parent are first determined, and one equal-fitness individual is then selected uniformly at random. NGD can be therefore considered as a random walk on the neutral neighborhood of equal-fitness offspring. A new population is formed in each generation with the selected parent from the previous population and the $\lambda$ bred offspring. An exemplification of the ($1+\lambda$)-EA variant used in CGP is provided in Algorithm~\ref{cgp_algorithm}. For the breeding procedure point mutation is predominantly used to exchange gene values of the genotype within the valid range by chance. The mutations triggered by this operator can alter the functionality of the phenotype as well as the connectivity depending on which type of gene is mutated. Genetic programs are mostly encoded with natural numbers in CGP that is commonly refereed to as integer-based or standard CGP. However, an alternative encoding model that is called real-valued CGP~\cite{DBLP:conf/gecco/CleggWM07} has been proposed. It uses real numbers to encode candidate programs with the intention to adapt intermediate recombination which is commonly used for real-valued representations in Genetic Algorithms~\cite{Eiben2015}. In contrast, integer-based CGP is predominantly used merely with mutation due to its long history of stagnation regarding the question about the effectiveness of recombination~\cite{DBLP:journals/gpem/Miller20}. Recent studies, however, have found that various recombination operators such as subgraph~\cite{DBLP:conf/eurogp/KalkreuthRD17}, block~\cite{DBLP:conf/eurogp/HusaK18} and discrete crossover~\cite{DBLP:conf/ppsn/Kalkreuth22} can be effectively used for various problems~\cite{10.1007/978-3-031-46221-4_3}.

\subsection{Problem Domains in GP}

GP gained significant popularity when Koza~\cite{koza1990genetic,Koza:1992:GPP:138936,koza1994genetic} applied his parse tree representation model to practically relevant problem domains, for instance, symbolic regression, algorithm construction, logic synthesis, or classification. In this section, we describe two popular representatives of the GP application scope in more detail which have a reputation for being major real-world application scopes for GP as well as for their relevance for benchmarking GP methods:

\subsubsection{Symbolic Regression}

Symbolic Regression (SR) is located in the broader taxonomy of regression analysis, where a symbolic search is performed in a space of mathematical compositions to obtain candidate functions that match the ideal input-output mapping of a given data set as closely as possible. Symbolic regression in the context of GP can therefore be considered a black-box problem domain. In general, SR by means of GP relates to the application of GP models to synthesize mathematical expressions that represent input-output mapping of the the unknown function as closely as possible. Quite recently, it has been proved SR to be a NP-hard problem, since it is not always possible to find the best fitting mathematical expression for a given data set in polynomial time~\cite{DBLP:journals/tmlr/VirgolinP22}.


\subsubsection{Logic Synthesis}

Logic synthesis~\cite{DBLP:books/daglib/0086041, hassoun2001logic} as tackled with GP comprises the synthesis of Boolean expressions that match input-output mappings of given Boolean functions. Boolean expressions are generally a way of formally expressing Boolean functions. LS as approached with GP predominantly addresses two major tasks located in the scope of this problem domain: 

\begin{enumerate}
\item Synthesis of a Boolean expression that matches the correct input-output mapping of a given Boolean function.
\item Optimization of a Boolean expression (i.e. reduction of complexity).
\end{enumerate} 


 Both tasks are carried out with respect to Boolean logic and algebra. Truth tables are a common way to represent Boolean functions and to describe their input-output mapping besides to representing them with algebraic expressions. Synthesis of Boolean expressions is typically approached by defining one or multiple respective optimization objectives. LS as an GP application area was greatly popularized by Koza when he started addressing LS by using his parse tree representation model~\cite{DBLP:conf/ijcai/Koza89, DBLP:conf/ppsn/Koza90, DBLP:conf/foga/Koza90}. Moreover, Koza utilized his approach to evolve expressions for Boolean functions such as digital multiplexers and parity since these functions can be represented as LISP S-expressions. However, digital circuits are often characterized by Boolean functions with multiple outputs such as digital adders or multipliers. This resulted in the predominant use of CGP for LS since its graph encoding model is well-equipped to represent such functions~\cite{DBLP:conf/ddecs/FiserSVS10, Sekanina2011}. A real-world application of the LS domain is the automatic design of digital circuits. 


\subsection{Modern C++}

C++ as a versatile and powerful programming language, has evolved significantly over the course of the last decade. Starting with the release of \texttt{C++11}~\cite{ISO:2012:III} and the subsequent versions C++14, C++17, and C++20, various new features and corresponding best practices have been introduced, allowing developers to write more efficient and maintainable programs. Moreover, features that are associated with modern C++ have noticeably changed the way code is written in C++ remarkably improved the safety and expressiveness of C++ and are provided in the C++ Standard Library. Some of the language features that shaped modern C++ are: \\

\begin{itemize}
\item \textbf{Template type deduction} \\
Templates are a feature that enables the use of generic types for functions and classes. Template type deduction therefore allows the creation of functions or classes that can be adapted to more than one type without re-implementing the code constructs for each type. In C++ this can be achieved using template parameters. \\
\item \textbf{Smart Pointers} \\
Smart Pointers provide a wrapper class around a raw pointer that have overloaded access operators such as  \texttt{*} and \texttt{->}. Smart pointer managed objects have similar appearance as regular (raw) pointers. However, smart pointers can be deallocated automatically, in contrast to raw pointers. Smart pointers are therefore used to ensure that programs are free of memory leaks and, in this way, simplify the dynamic memory allocation in C++ while maintaining efficiency. \\
\item \textbf{Lambda Expressions:} \\
A concise method for defining inline functions or function objects is to use lambda expressions, especially when working with algorithms or when a function is used as parameter. Lambdas can make the code more readable by allowing more direct expressions of intentions, since they do not require explicit function declarations. Lambdas are, therefore, also called anonymous functions.  \\
\item \textbf{Constexpr} \\
The primary intention behind constant expression is to enable performance improvement of programs by doing computations at compile time rather than runtime. C++11 introduced the keyword \texttt{constexpr}, which declares that it is possible to evaluate the value of a certain function or variable at compile time. \\
\item \textbf{Concurrency} \\
Concurrency support was initially introduced in \texttt{C++11} in order to boost program efficiency and allow multitasking. The Concurrency Support Library of C++ provides support for threads, atomic operations, mutual exclusion, and condition variables. Although concurrency enables multitasking, it does not necessarily mean that the desired tasks are executed simultaneously but are more approached by efficient switching between tasks.
\end{itemize} 

\section{Existing CGP Implementations}
\label{sec:implementations}
This section reviews existing implementations of CGP that are later considered for a comparison to \texttt{CGP++}. Some implementations have already been addressed in Miller's review on the state and future of CGP~\cite{DBLP:journals/gpem/Miller20}. However, with the intention to complete the picture further and to allow a more comprehensive comparison, we consider additional implementations and address their key features and purpose briefly. Despite the fact that the resource for Julian Miller's C reference implementation went down a while ago, a modified version still exists and is publicly available\footnote{\url{http://github.com/paul-kaufmann/cgp/}}. The modified version has been adapted for hyperparameter tuning experiments and search performance evaluations across several methods suitable for combinatorial optimization and combinational synthesis~\cite{DBLP:conf/cec/KaufmannK20, 
DBLP:conf/gecco/SottoKAKB20}. The implementation has therefore been additionally equipped with search algorithms such as simulated annealing. Another C implementation is the \texttt{CGP-Library}\footnote{\url{http://www.cgplibrary.co.uk/}} published by Turner and Miller~\cite{DBLP:journals/gpem/TurnerM15}. It supports standard CGP as well as the recurrent CGP~\cite{DBLP:conf/ppsn/TurnerM14} variant and provides the functionality for evolving artificial neural networks~\cite{DBLP:journals/ijon/KhanAKM13, DBLP:conf/gecco/TurnerM13}. The popular Java-based Evolutionary Computation Research System (ECJ)~\cite{DBLP:conf/gecco/ScottL19, DBLP:conf/gecco/Luke17} provides a CGP contrib package that supports integer-based CGP as well as real-valued CGP. Moreover, the ECJ CGP contrib package covers functionality, data and benchmarks for applications such as logic synthesis, classification and symbolic regression. Recently, a set of implementations of various advanced genetic operators has been added to the repository. The \texttt{CGP Toolbox}\footnote{\url{http://www.fit.vutbr.cz/~vasicek/cgp/}} is a framework that primarily focuses on LS addressed with CGP and has been proposed by Vasicek and Sekanina~\cite{vasicek:2012:EuroGP}. It is shipped in four different versions that, in each case, support LS or SR for either 32 or 64 bit architecture and enables efficient phenotype evaluation based on machine code vectorization. A CGP toolbox for Matlab that focuses on audio and image processing called~\texttt{CGP4Matlab} has been proposed by Miragaia et al. ~\cite{10.1007/978-3-319-77538-8_31} which was used to apply CGP to the problem of pitch estimation. \texttt{HAL-CGP}\footnote{\url{http://github.com/Happy-Algorithms-League/hal-cgp}} is a pure Python implementation of CGP designed to target applications that are characterized by computationally expensive fitness evaluations~\cite{schmidt_2020_3889163}. ~\texttt{CartesianGP.jl}\footnote{\url{http://github.com/um-tech-evolution/CartesianGP.jl}} is a library for using CGP in Julia. 
However, according to the authors, the code should be considered \textit{pre-alpha} at the moment.

\section{The Proposed Implementation}
\label{sec:proposal}

\subsection{General Motivation and Philosophy}

Since Miller officially proposed CGP, increasing development has taken place over the course of the past two decades in the relatively young field of graph-based GP by proposing new representation variants, promising forms of crossover, mutation and search algorithms, as well as benchmarks. 
With the proposal of \texttt{CGP++}, we think that our proposed implementation can address the following aspects to enhance the following points in the field of CGP: 
\begin{itemize}
    \item Maintaining accessibility for the use of CGP by extending the ecosystem of existing CGP implementations
   \item Improving the interpretability of sophisticated  methods by providing a comprehensible architecture.
    \item Facilitating reproducibility of existing results by supporting benchmarking frameworks.  
\end{itemize}

The fundamental philosophy behind \texttt{CGP++} is to utilize aspects of modern C++ that have been described in Section~\ref{sec:related_work} to implement features and properties that are provided by state-of-the-art (SOTA) heuristic frameworks. In the following subsections, we will address the key features and properties of \texttt{CGP++} and share some details about the respective implementation details of that we used from modern C++. 

\subsection{Key Features and Properties} 

\subsubsection{Object-oriented and Generic Design}
\texttt{CGP++} pursues an object-oriented and generic design to maintain an interpretable and reusable architecture for fundamental as well as sophisticated functionality, with the intention to assist further implementations of new techniques and the corresponding extension of the underlying architecture. 

\subsubsection{Advanced Genetic Variation}

Since CGP has been predominantly used without recombination in the past, most implementations only support CGP in the standard mutation-only fashion. However, since recent work proposed new recombination operators and demonstrated the effectiveness for various problems~\cite{10.1007/978-3-031-46221-4_3}, block ~\cite{DBLP:conf/eurogp/HusaK18} and discrete recombination~\cite{DBLP:conf/ppsn/Kalkreuth22} have been integrated into \texttt{CGP++}. Besides to the (1+$\lambda$)-EA variant used in CGP that has been exemplified in Algorithm~\ref{cgp_algorithm}, an implementation of a ($\mu$+$\lambda$)-EA is provided to allow the recombination-based use of CGP. Furthermore, since recent work demonstrated that the consecutive execution of multiple mutation operators can benefit the search performance of CGP~\cite{DBLP:conf/ijcci/Kalkreuth22, 10.1145/3520304.3529065}, \texttt{CGP++} therefore supports mutation pipelining and provides advanced mutations such as \textit{inversion} and  \textit{duplication}.

\subsubsection{Benchmarking}

\texttt{CGP++} provides an interface to the benchmarks that have been recently proposed for the General Boolean Function Benchmark Suite (GBFS) which provides a diverse set of LS problems for GP~\cite{10.1145/3594805.3607131}. The provided \texttt{PLU} files contain compressed truth tables that can be used to set up the corresponding black-box problem. Moreover, \texttt{CGP++} also provides a dataset generator and a set of objective functions for the SR benchmarks that have been proposed by McDermott et al.~\cite{DBLP:conf/gecco/McDermottWLMCVJKHJO12} in the framework of the first review on benchmarking standards in GP.

\subsubsection{Hyperparameter Configuration}

Hyperparameters related to the CGP functionality can be configured by using either a provided command-line interface or a parameter file, offering a flexible approach that can be used to apply \texttt{CGP++} to contemporary frameworks for hyperparameter tuning such as \texttt{irace}~\cite{LOPEZIBANEZ201643}. 

\subsubsection{Checkpointing}

To ensure caching of intermediate search results that have been obtained over the course of the search process, \texttt{CGP++} supports the creation of checkpoints that are automatically written during a run. The created checkpoint file can be used to resume a run in the case that it has been disrupted. 

\subsection{Implementation Details and Challenges}

\subsubsection{Generic Template}

The generic template of \texttt{CGP++} can be formally described as a tuple $\mathcal{T} = (\mathcal{E},~\mathcal{G},~\mathcal{F})$ where $\mathcal{E}$ defines the evaluation type, $\mathcal{G}$ the type of the CGP genome and $\mathcal{F}$ the type of fitness. \texttt{CGP++}'s generic approach is achieved by using C++ class templates. The respective data types can be configured via \texttt{typedef}. To restrict the data type of certain template classes such as the type of the genome, we use \texttt{constexpr} to evaluate the defined template type at compile time.   

\subsubsection{Smart Memory Management}

\texttt{CGP++} utilizes two types of smart pointers: \texttt{std::unique\_ptr} and \texttt{std::shared\_ptr} to provide safe memory allocation as well as efficient passing of objects, containers and data to functions and classes. In our implementation, \texttt{std::shared\_ptr} is used for shared ownership of objects among instances of different classes. In contrast, \texttt{std::unique\_ptr} is used in cases where single or exclusive ownership of a resource is desired.

\subsubsection{Memorization}

Memorization is used to speed up genotype-phenotype decoding by caching the immediate results of function nodes and consequently preventing reevaluating already computed results. The node-value mapping are stored by using \texttt{std::map} during the decoding routine. 

\subsubsection{Concurrency}

Besides to consecutive evaluation of individuals, \texttt{CGP++} takes the first step towards concurrency by providing concurrent evaluation of the population. For this purpose, the population is divided into chunks of individuals whose number is defined by the number of evaluation threads that can be set in the configuration. The chunks are then evaluated within several instances of \texttt{std::thread}. \texttt{CGP++} supports deep cloning of problem instances to create the corresponding thread pool. The pool of threads are synchronized after evaluation via \texttt{std::thread::join}. However, the genotype-phenotype mapping of CGP and the corresponding requirement of a decoding procedure poses a bottleneck for the use of concurrency. At this time, we have to limit the concurrency feature in \texttt{CGP++} for parts of the decoding and evaluation procedure of the genotype by using mutual exclusion via \texttt{std::mutex}. We will address potential solutions for this issue in the discussion. 

\subsection{Resources}

The source code of \texttt{CGP++} and a user guide are publicly available in our GitHub repository\footnote{\url{http://github.com/RomanKalkreuth/cgp-plusplus}}. 

\section{Architecture and Workflow Overview}
\begin{figure*} 
  \includegraphics[scale=0.67]{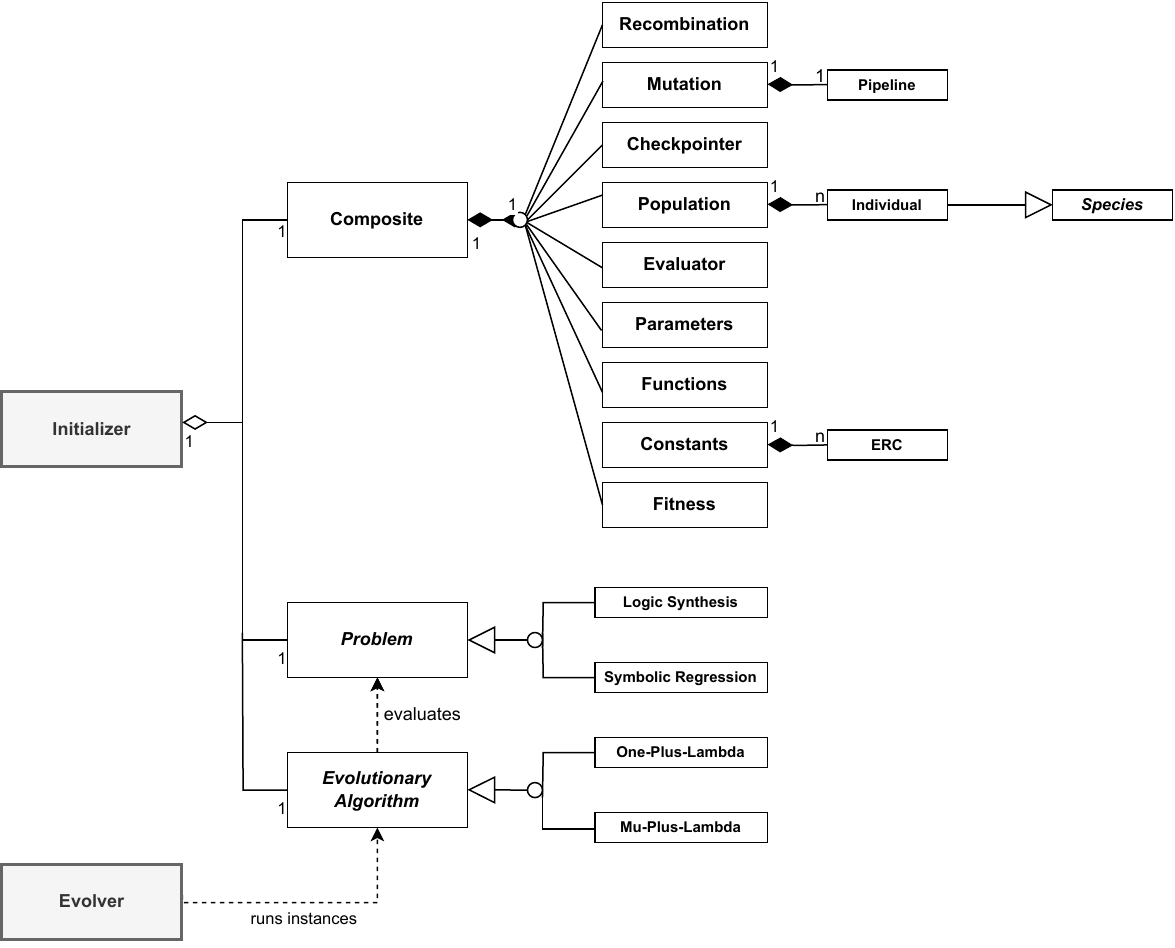}
  \caption{Illustration of the high-level architecture of \texttt{CGP++} which uses abstraction and inheritance to facilitates reusability and flexibility for further extensions. Smart pointers are used to provide safe and efficient  access to elements across the framework.}
  \label{img:architecture}
\end{figure*}

\label{sec:overview}
\subsection{Fundamental Architecture}
Abstraction and inheritance depict fundamental pillars of the top level architecture of-\texttt{CGP++} to enable a high degree of reusability of its core functionality. Figure~\ref{img:architecture} provides a high-level class diagram that covers its key architecture elements. The \textit{Initializer} is designed to instantiate the core elements for the heuristic search performed by CGP, such as the selected search algorithm and defined problem. To bundle essential sub-elements for the heuristic search process, a \textit{composite} is initialized, which can be accessed by other core elements such as the problem and algorithm instances. The composite includes crucial elements and features for the GP search process, such as the population, breeding execution frameworks for crossover and mutation, function and terminal (constants) sets but also backbone elements such as hyperparameter interfaces as well as checkpointing. \texttt{CGP++} supports the generation of ephemeral random constants (ERC) to create the terminal set, which, together with the function set, is an integral part of the \textit{Composite}. After initialization, the \textit{Evolver} executes the considered number of instances (jobs) and reports final as well as immediate results via command line and output file. The high-level architecture of \texttt{CGP++} is fundamentally inspired by ECJ~\cite{DBLP:conf/gecco/ScottL19, DBLP:conf/gecco/Luke17}. \\

\subsection{Top-level Workflow}
\begin{figure}
  \includegraphics[scale=0.54]{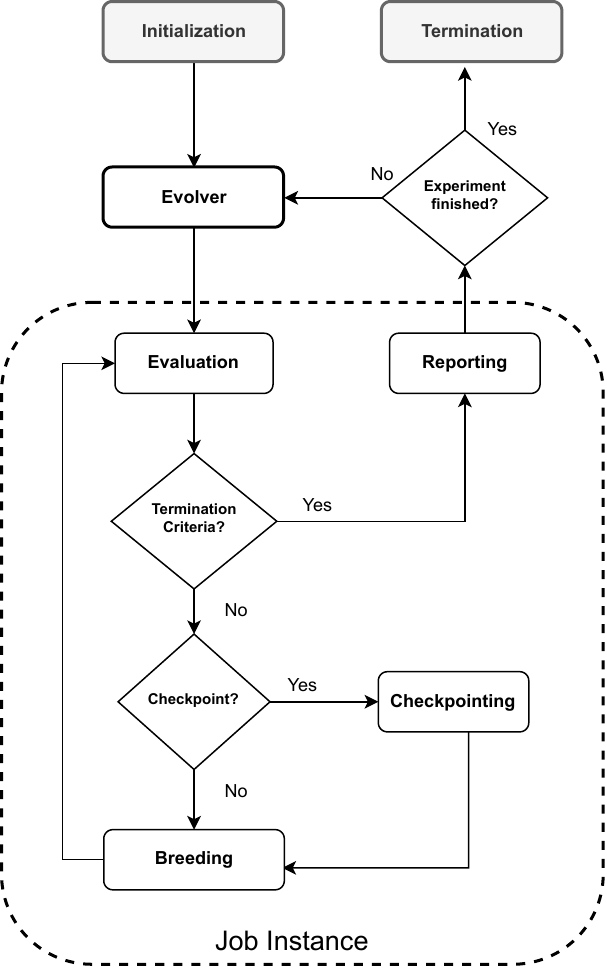}
  \caption{Overview of \texttt{CGP++}'s top-level workflow, addressing the execution of run instances as well as the main workflow that is executed within an instance.}
  \label{img:workflow}
\end{figure}

The top-level workflow of \texttt{CGP++} is shown in Figure~\ref{img:workflow} that illustrates the interplay between the elements that have been described related to the fundamental architecture. \texttt{CGP++} can be used to run experiments that require several instances to ensure statistical validity. The \textit{Evolver} therefore supports the execution of consecutive jobs, whose numbers can be configured via the parameter interface. The workflow within the framework of a job instance maps the typical workflow of the adapted (1+$\lambda$) and ($\mu$+$\lambda$) strategies. To facilitate the integration of other types of evolutionary algorithm, we provide an abstract base class that can be used as a design pattern. \\

\subsection{Concurrent Evaluation}
\begin{figure}
  \includegraphics[scale=0.57]{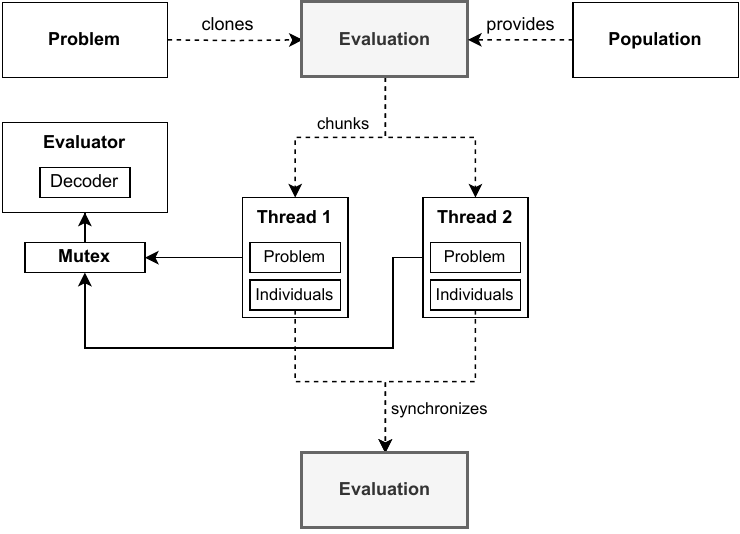}
  \caption{Illustration of the architecture for concurrent evaluation. In this example, the thread pool is simplified with two threads to give a structured overview. However,  in general, \texttt{CGP++} is capable of instantiating and executing more than two threads.}
  \label{img:evaluation}
\end{figure}

The concurrent evaluation architecture is illustrated in Figure~\ref{img:evaluation}. When concurrent evaluation of the individuals is used, the evaluation procedure forks and joins a thread pool and each thread is equipped with a chunk of individuals as well as a deep copy of the problem instance. The respective functionalities such as deep cloning are provided in the \textit{Population} and \textit{Problem} classes. Since the \textit{Decoder} is the shared resource in this framework, its access is maintained via mutual exclusion, as already mentioned in the previous section.  

\subsection{Checkpointing}
\begin{figure}
  \includegraphics[scale=0.58]{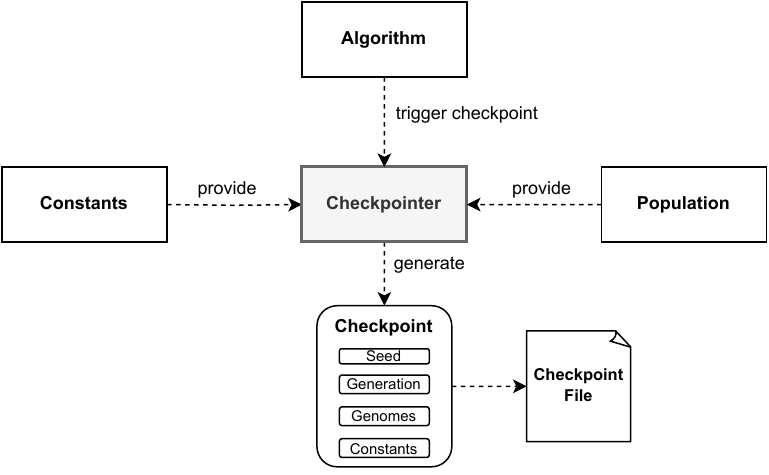}
  \caption{Illustration of the generation of a checkpoint that is generated with the chromosomes obtained from the population, the constants (ERC), the seed of the random generator, and the generation number. The checkpoint is written to an output file that can be used to resume a run. }
  \label{img:checkpoint}
\end{figure}

The generation of a checkpoint is shown in Figure~\ref{img:checkpoint}. For a checkpoint, we consider the random seed, generation number, genomes of the population, and the constants. These attributes are obtained from the respective instances and are then considered as a checkpoint instance that is written to a checkpoint file. Instances can be resumed by using the same configuration as the aborted run and passing the checkpoint file to \texttt{CGP++}. When a checkpoint file is detected, it is loaded by the \textit{Checkpointer} and the run instance is resumed at the given generation number. 

\section{Implementation Comparison}
\label{sec:comparison}
We considered various features for our comparison that can be found in modern metaheuristics toolkits. These include programming design, generic properties, checkpointing, variation pipelining, concurrency, and the existence of a parameter interface that can be used for hyperparameter tuning. We consider vectorization as a feature for our comparison, since it can be seen as a competitive feature to concurrency in CGP. With respect to recent findings about the role of crossover in CGP, we also consider this feature in our comparison. Table~\ref{tab:comparison} shows the result of our comparison, and it is visible that the the supported features of our implementation are on the level of a modern metaheuristics toolkit such as ECJ. Please note that the evaluation of Julian Miller's reference implementation is based on its modified version. The ECJ CGP contrib package offers a wide range of features that are derived from the underlying ECJ framework. For the other implementations, we notice that the number of features is quite limited. However, the \texttt{CGP Toolbox} supports vectorization, which is currently not supported by \texttt{CGP++}. A finding that we will address in the following discussion. \texttt{CGP4Matlab} and \texttt{HAL-CGP} use programming languages that pursue dynamic typing concepts. However, we do not consider these concepts as generic, since generic programming aims at enabling data type independence while maintaining compile-time type safety. To our best knowledge, this is not covered by default by the dynamic typing concepts of these languages, and the generic extensions and features are currently not used for the respective implementations. 

\begin{center}
\begin{table}
\hspace{-8mm}
\scalebox{0.75}{
\begin{tabular}{r|cc|cccccccc}
&
\rot{Language} &
\rot{Paradigm} &
\rot{Generic Design} &
\rot{Checkpointing} &
\rot{Concurrency} &
\rot{Pipelining} &
\rot{Crossover} &
\rot{Vectorization} &
\rot{Parameter Interface} \\ \hline
CGP++ & C++ & O & \ding{51}  & \ding{51} & \ding{51} & \ding{51} & \ding{51} & - & \ding{51}\\ 
ECJ (CGP contrib)  & Java & O & - & \ding{51} & \ding{51} & \ding{51} &  \ding{51} & - &  \ding{51} \\
Reference (Miller) & C &  P & - &  -& - & - & - & -&\ding{51} \\
CGP-Library & C++ &  P  & - & - & \ding{51} &  - & - & - & -\\ 
CGP Toolbox & C++ & P & - & - & - & - & - & \ding{51} & \ding{51} \\ 
CGP4Matlab & Matlab & O & - & - & - & - & - & - & -   \\
HAL-CGP & Python & O & - & - & - & - & - & - & -\\ 
CartesianGP.jl & Python & P & \ding{51} & - & - & - & - & - & - \\ \hline
\end{tabular}
}\\
\textbf{O}: object-oriented \quad \textbf{P}: procedural 
\caption{Results of our implementation comparison that considered different general features of modern heuristic toolkits but also CGP related aspects.}
\label{tab:comparison}
\end{table}
\end{center}

\section{Discussion and Future Work}

The primary intention of this work is to propose the first version of a modern implementation of CGP in the C++ programming language. Another intention behind our work is to propose and establish a flexible and reusable architecture that can facilitate and simplify the implementation of further extensions. We therefore deliberately chose C++ over Rust which would also have been a suitable option. However, we consider Rust more a procedural and functional-oriented programming language rather than a strong object-oriented one. Moreover, since we also focus on interpretability of CGP methodologies implemented in \texttt{CGP++}, we find C++ code more approachable than Rust code for this purpose. 

We want to stress that with \texttt{CGP++} we do not intend to propose a framework that we generally consider superior over other implementations. With our contribution, we more intend to extend the already-existing ecosystem of CGP implementations. We acknowledge that every programming language has its own specific characteristics, and each implementation has its own purpose and philosophy. However, we feel that most CGP implementations fall short of providing features that can facilitate the discovery of novel applications and the integration of new techniques. Moreover, since the majority of the implementations that we considered for our comparison follow the procedural programming paradigm, we think that there is a need for frameworks that can facilitate the implementation and maintenance of larger and more complex methods that have been proposed for CGP. 
Another point that should be discussed is related to how CGP is effectively used and how this is related to future work on \texttt{CGP++}. CGP has a reputation for being effectively used with relatively small population sizes due to early experiments with Boolean functions~\cite{10.5555/2934046.2934074, DBLP:journals/tec/MillerS06}. However, recent studies on the parametrization of CGP demonstrated that CGP can be also effectively used with large population sizes in the SR domain~\cite{DBLP:conf/ki/KaufmannK17}. In contrast, these studies also demonstrated that CGP performs best in the LS domain when a (1+$\lambda$)-ES with a very small population size is used. Moreover, very recent work found the ($1+1$)-ES to be the best choice for the evaluation of the General Boolean Function Benchmark Suite (GBFS)~\cite{10.1145/3594805.3607131} for LS. Based on these findings, we think that at least two modalities in CGP have to be considered for future work. Therefore, we would like to address the following point as natural next steps for \texttt{CGP++}:

\label{sec:discussion}
\subsection{Concurrency}

Since we already raised the issue of using concurrency for CGP, we would like to discuss how the corresponding challenges could be tackled in the future. In the first place, this would imply extending the thread pool design by using multiple evaluator instances. However, we have to stress here that related elements such as the CGP decoder have to be multiplied, and this would lead away from the idea of using a lightweight thread pool inside \texttt{CGP++}. Another idea would be to consider an alternative concurrency design pattern that could enable highly concurrent use and implements aspects from parallel dynamic programming~\cite{STIVALA2010839}. Currently, \texttt{CGP++} only supports concurrency for the evaluation process. Therefore, another contribution would be to enable breeding concurrency. In general, despite the highlighted challenges, we think it is worthy to explore whether concurrency could be effectively used for problems where CGP seems to work well when large population sizes are being considered. 

\subsection{Vectorization}

The use of vector operations by using related extended SIMD instructions is another feature that we would like to consider for future work. Vectorization with machine code that has been generated from CGP primitives and which contains SIMD instructions has been successfully used to speed up the CGP evaluation procedure~\cite{vasicek:2012:EuroGP}. The used SSE/SSE2 SIMD instruction calls operated with 128-bit vectors in that case. However, providing such a feature from today's perspective could also enable support for contemporary instruction sets such as Advanced Vector Extensions (i.e. AVX-2 or AVX-512). In view of the fact that the ($1+1$)-ES has been found to be the best choice on GBFS, vectorization could be used to provide a way to use CGP effectively in a consecutive fashion. 

\subsection{Towards a Modern General GP Toolkit}

Even if this paper proposes the first version of \texttt{CGP++}, we do not only see it as a implementation of CGP but also as a blueprint that can shape the way towards a modern and general GP toolkit that allows the use of multiple GP variants in a flexible and effective way. Therefore, we consider extending \texttt{CGP++} to \texttt{GP++} in the future that can benefit the GP domain across different representations with the contributing factors that we intend to achieve with the proposal of our implementation. As a first step towards that goal, we plan to integrate tree-based and linear GP as popular representatives of the GP domain. 

\section{Conclusion}
\label{sec:conclusions}

In this paper we presented the first version of a modern C++ implementation of Cartesian Genetic Programming, which closes a major gap in the framework of existing implementations. Our implementation provides key features and characteristics of modern heuristics frameworks. Our proposed implementation offers a genetic design and provides a reusable architecture that can facilitate the discovery of new problem domains and the integration of new methods for CGP. Equipped with interfaces and generators for benchmarking in Logic Synthesis and Symbolic Regression, \texttt{CGP++}, also provides a framework that can be used for the reproducibility of existing results.

\paragraph{\textbf{Acknowledgments}}
\texttt{CGP++} is dedicated to Dr. Julian Francis Miller (1955 - 2022), who as the founder of Cartesian Genetic Programming devoted a large part of his scientific life to its proposal, development and analysis. With the introduction of \texttt{CGP++} we pay tribute to Julian's pioneering effort in the field of graph-based GP and acknowledge his lifework. The project was financially supported by ANR project HQI ANR-22-PNCQ-0002.

\clearpage
\appendix 
\section{Appendix}
\begin{table}[H]
\caption{Notation}
\label{tab:notations}
\scalebox{1.0}{
\begin{adjustbox}{center}
    \begin{tabular}{c|c}
    \toprule
    Symbol          & Definition   \\ \hline
    $\mathfrak{P}$ & Genetic program \\ 
    $\Psi$	& Phenotype \\ 
    $\mathfrak{F}$ & Set of functions \\ 
    $\mathfrak{T}$ & Set of terminals \\ 
    $\mathfrak{E}$ & Set of edges \\ 
    $\mathfrak{N_\textnormal{i}}$ & Set of input nodes \\ 
    $\mathfrak{N_\textnormal{f}}$ & Set of function nodes \\ 
    $\mathfrak{N_\textnormal{o}}$ & Set of output nodes \\ 
    $\mathcal{T}$ & Generic template \\
    $\mathcal{E}$ & Evaluation type \\
    $\mathcal{F}$ & Fitness type \\
    $\mathcal{G}$ & Genome type\\
    \bottomrule
    \end{tabular}
\end{adjustbox}
}
\end{table}

\bibliographystyle{ACM-Reference-Format}
\bibliography{bib/gecco24}

\end{document}